\documentclass[10pt,twocolumn,letterpaper]{article}

\usepackage[arxiv]{cvpr} 

\usepackage{textcomp}
\usepackage{amsmath}
\usepackage{amssymb}
\usepackage{algorithm}      
\usepackage{algorithmic}    
\usepackage{soul}
\usepackage{multirow}    
\usepackage{array}       
\usepackage{xcolor}      
\usepackage{booktabs}    

\usepackage[table,xcdraw]{xcolor}
\definecolor{lightgray}{gray}{0.95}
\definecolor{midgray}{gray}{0.85}
\definecolor{HLGreen}{rgb}{0.78,0.95,0.78}
\definecolor{HLRed}{rgb}{0.9725,0.8431,0.8549}   
\definecolor{benchblue}{RGB}{58, 122, 252}

\definecolor{bestcolor}{RGB}{219, 208, 237}
\definecolor{secondcolor}{RGB}{241, 237, 248}
\definecolor{thirdcolor}{RGB}{211, 222, 190}
\definecolor{line-blue}{RGB}{243, 248, 252}

\definecolor{cvprblue}{rgb}{0.21,0.49,0.74}
\usepackage[pagebackref,breaklinks,colorlinks,allcolors=cvprblue]{hyperref}
\usepackage{tikz}

\def\confName{CVPR}
\def\confYear{2026}

\newcommand{\methodname}{ReMoT}
\newcommand{\method}{\texttt{\methodname}\xspace}
\title{ReMoT: Reinforcement Learning with Motion Contrast Triplets}

\author{
Cong Wan$^1$\quad Zeyu Guo$^1$\quad Jiangyang Li$^1$\quad Songlin Dong$^2$$\thanks{Corresponding author: Songlin Dong and Zhiheng Ma }$\quad Yifan Bai$^3$\quad Lin Peng$^1$\\[2pt]
Zhiheng Ma$^{2*}$\quad Yihong Gong$^1$\\[4pt]
$^1$Xi'an Jiaotong University\\
$^2$Shenzhen University of Advanced Technology\\
$^3$DAMO Academy, Alibaba Group\\[2pt]
\small \texttt{wancong@stu.xjtu.edu.cn\quad dongsl@suat-sz.edu.cn}
}

\begin{document}
\maketitle

\begin{abstract}
We present \method, a unified training paradigm to systematically address the fundamental shortcomings of VLMs in spatio-temporal consistency—a critical failure point in navigation, robotics, and autonomous driving. \method integrates two core components: (i) A rule-based automatic framework that generates \method-16K, a large-scale (16.5K triplets) motion-contrast dataset derived from video meta-annotations, surpassing costly manual or model-based generation. (ii) Group Relative Policy Optimization, which we empirically validate, yields optimal performance and data efficiency for learning this contrastive reasoning, far exceeding standard Supervised Fine-Tuning.  We also construct the \textbf{first} benchmark for \textbf{fine-grained} motion contrast triplets to measure a VLM's discrimination of subtle motion attributes (e.g., opposing directions). The resulting model achieves SOTA performance on our new benchmark and multiple standard VLM benchmarks, culminating in a remarkable \textbf{25.1\%} performance leap on spatio-temporal reasoning tasks.
\end{abstract}    
\section{Introduction}

\begin{figure*}[t]
  \centering
  \includegraphics[width=\linewidth]{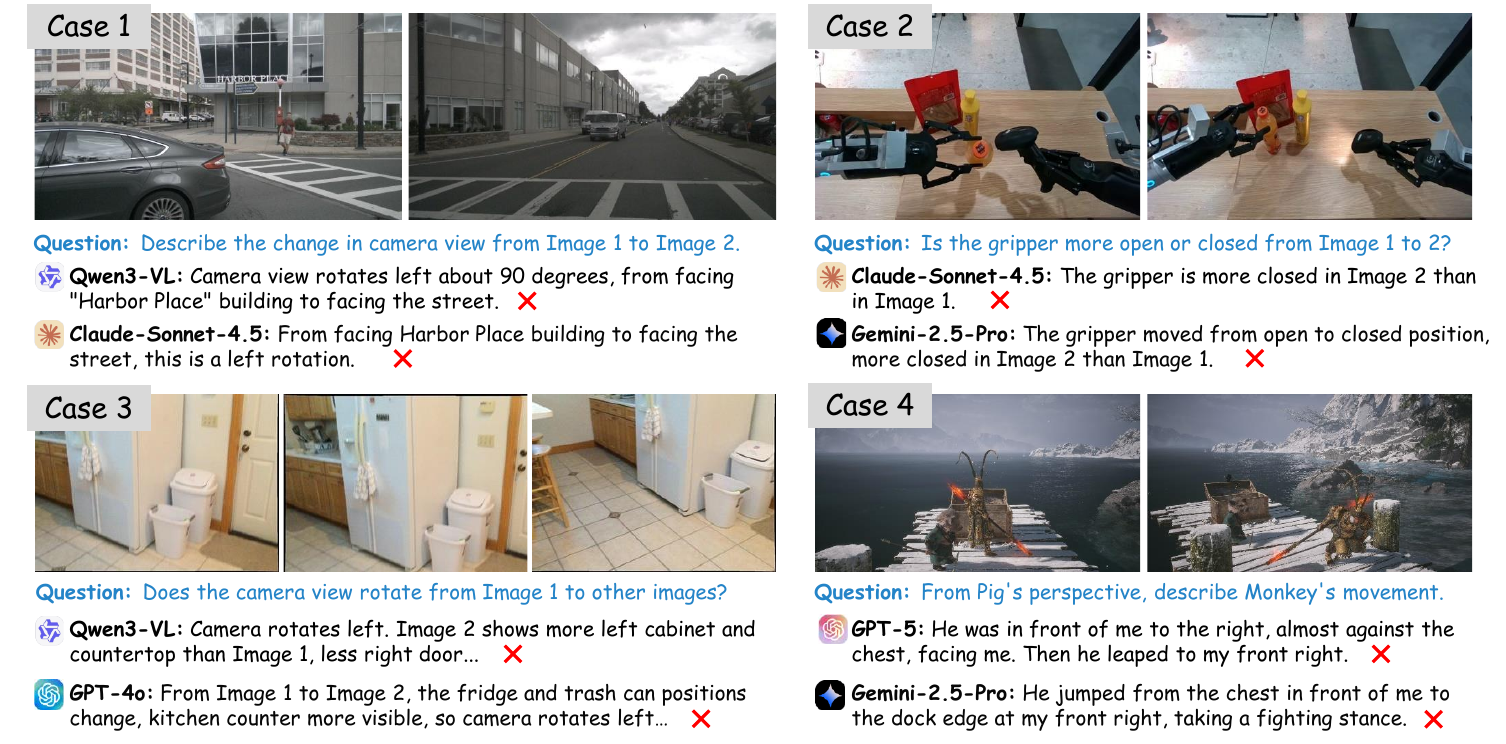}
\caption{
\textbf{Common failure modes of large Vision–Language Models (VLMs) in spatio–temporal reasoning.}
The figure presents four multi‑image examples drawn from navigation, robotic manipulation, indoor exploration, and game simulation scenarios.  
Each example provides multiple related images and a question about their spatial or temporal relationship.  
Recent VLMs (GPT‑4o, Claude‑Sonnet‑4.5, Gemini‑2.5‑Pro, Qwen3‑VL) give incorrect responses—such as reversing camera rotation, misjudging object openness, or confusing character motion—which are indicated by red crosses.  
The errors illustrate that current VLMs struggle to reason consistently about spatial correspondence and physical change across multiple views.
}
  \label{fig:pilot}
\end{figure*}

Vision–Language Models (VLMs) have rapidly evolved into general-purpose perception systems that unify visual understanding and language reasoning~\cite{o1, deepseekr1, team2025kimi1.5, seedthinker}. They are increasingly being deployed in critical domains involving physical world interaction, such as AIGC~\cite{song2025layertracer, song2025omniconsistency, song2024diffsim, song2024processpainter, wan2024grid,unireal,stepxedit,qwenimage}, embodied intelligence~\cite{openvla,pi0,gr3,worldvla,robix}, 
and autonomous driving~\cite{vlmad,bai3d}. These tasks fundamentally require models to move beyond static, single-frame perception and instead reason about how physical scenes evolve in space and time, for example, tracking 3D object motion~\cite{spatialvlm,vlm4d}, interpreting dynamic changes across consecutive frames~\cite{vlmad}, or inferring global orientation~\cite{vlmaps,navid}. However, current mainstream VLMs~\cite{qwen2.5vl,chen2024internvl,llava,gpt4o,gpt4} exhibit significant limitations in this regard: While they excel at aligning visual semantics, they suffer from fundamental deficiencies in the core capability of ensuring spatial-temporal consistency.

Our empirical analysis, spanning diverse scenarios such as navigation, robotic manipulation, indoor exploration, and game simulation, reveals the pervasiveness of this shortcoming. As illustrated in Fig.~\ref{fig:pilot}, even top-tier general-purpose large models, including Qwen3-VL, Claude-Sonnet-4.5, and the GPT series, due to lacking a robust understanding of spatial-physical regularities, frequently confound camera rotation with real object motion (Case 1,3), misinterpret gripper status (Case 2), and erroneously infer the direction of character movement (Case 4). However, existing methods, including architectural modifications~\cite{SpatialMLLM,vlm3r} or data augmentation~\cite{mico,spatialssrl,simsv,noisyrollout}, are largely limited to piecemeal fixes. They fail to provide a systematic solution that addresses this fundamental limitation across the paradigms of data, training, and evaluation.

To systematically address this fundamental deficiency, our solution unfolds across three dimensions:

\noindent\textbf{Data:} Existing VLM training data predominantly relies on static image-text pairs, lacking explicit modeling of fine-grained motion attributes and proving insufficient for learning fine-grained spatio-temporal reasoning. To address this, we construct \method-16K, a large-scale motion-contrast triplet dataset to explicitly model fine-grained inter-frame motion attributes (e.g., ``camera rotates left" vs. ``camera rotates right"). 
For large-scale, high-quality generation, we propose a \textbf{multi-expert collaborative pipeline}. This pipeline orchestrates specialized components operating on structured meta-annotations from video datasets (\textit{e.g.}, camera pose matrices, robot action logs): motion estimation experts extract precise geometric and physical motion properties, triplet construction experts synthesize hard negatives via property-conditioned transformations, and VQA formulation experts design multi-perspective reasoning chains. This expert-driven approach significantly surpasses the scale and consistency of manual annotation or direct VLM-based generation (which suffers from 55\% format errors and limited valid outputs).

\noindent\textbf{Training:} 
Based on the constructed motion-contrast data, we systematically investigate the effectiveness of various optimization paradigms. Beyond standard Supervised Fine-Tuning (SFT), we explore Reinforcement Learning with composite rewards (combining task accuracy, reasoning compactness, and logical consistency within the GRPO framework), as well as hybrid strategies including sequential (SFT$\rightarrow$GRPO) and alternating (SFT$\leftrightarrow$GRPO) integration schedules.

\noindent\textbf{Benchmark:} For evaluation, we constructed a benchmark focusing on fine-grained motion contrast. Unlike existing benchmarks \cite{mmsibench,vsibench}, we systematically design sample pairs that are visually highly similar yet possess opposing motion attributes (e.g., ``translate left" vs. ``translate right" in ego-motion). Through diverse question formats (e.g., single-choice, multiple-choice, fill-in-the-blank), the model is required to reason about complex inter-frame dynamics, such as object motion, camera rotation, and physical continuity, rather than static recognition. The benchmark encompasses tasks including embodied navigation, robotic manipulation, and simulated game scenarios.

Our systematic exploration demonstrates that \textbf{\method}, a complete training paradigm integrating ``rule-driven motion-contrast data construction'' with ``GRPO optimization", provides a scalable and efficient solution for enhancing the spatio-temporal reasoning capabilities of VLMs. Compared to manual annotation or SFT methods, \textbf{\method} exhibits significant advantages in data generation efficiency, training sample utilization, and final performance, achieving SOTA on our new benchmark and multiple standard VLM benchmarks, and culminating in a remarkable \textbf{25.1\%} performance leap on spatio-temporal reasoning tasks.

\section{Related Work}
\label{sec:rel_work}

\noindent\textbf{VLM Spatio-Temporal Reasoning}
Vision-language models~\cite{gpt4o,gemini,qwen2.5vl} have advanced significantly, yet 
benchmarks~\cite{vlm2bench,vsibench,mmsibench,disheng2024thinking} consistently 
reveal failures in spatio-temporal reasoning: VLMs perform adequately on egocentric spatial judgments but fail at allocentric viewpoint reasoning 
and cross-viewpoint comprehension~\cite{mico,li2025viewspatial}. 
Works~\cite{spatialvlm,robospatial} generate static 3D relationships through depth lifting 
and scene graphs, scaling to billions of QA 
pairs. 
Temporal reasoning methods employ time-aware 
encoders~\cite{kim2024temporalvlm}, disentangled attention~\cite{MASH-VLM}, or textual temporal 
transfer~\cite{t3}, yet struggle with temporal concepts and rely on temporal proximity rather than 
contrastive motion semantics. 3D/4D approaches integrate geometry priors via 
reconstructive tuning~\cite{vlm3r} or world model latents~\cite{xu2024dyva,sun2024uni4d}, 
but static encoders weaken spatial grounding with sequential cues~\cite{canworld}
and require costly depth sensors.
However, these methods tackle isolated aspects (data scale, architectural design, 
evaluation metrics) without a unified framework addressing motion understanding holistically.
Our work addresses this through: (i) motion-contrast triplet 
construction from metadata, (ii) hybrid SFT-GRPO training for reasoning consistency, 
(iii) allocentric evaluation protocols capturing fine-grained motion semantics.

\noindent\textbf{Data Construction for Motion Supervision}
Existing training data for motion understanding suffers from insufficient granularity: 
video-text datasets provide only coarse clip-level captions~\cite{Bain21,grauman2022ego4d}, 
while embodied datasets offer structured metadata (poses, actions), but limited semantic 
annotations~\cite{savva2019habitat,dai2017scannet}. Recent advances explore self-supervised 
learning~\cite{spatialssrl,vjepa2}, contrastive training~\cite{mico,tclr}, 
and temporal modeling~\cite{only}, yet they either focus on spatial geometry over 
motion semantics~\cite{spatialssrl}, require massive unlabeled video~\cite{vjepa2}, 
or use temporal proximity as a weak motion proxy~\cite{tclr}—failing to explicitly 
model contrastive motion attributes.
In contrast, our motion-contrast triplets explicitly encode directional semantics 
by pairing each anchor with a motion-aligned positive (e.g., ``left rotation by 20°") 
and a motion-opposing hard negative synthesized from 
metadata. This enables scalable supervision for fine-grained motion discrimination 
unavailable in proximity-based or reconstruction-based paradigms.



\noindent\textbf{Training Paradigms for Reasoning Enhancement}
Vision-Language Models (VLMs) for spatio-temporal reasoning typically employ supervised fine-tuning (SFT) with techniques like LoRA~\cite{hu2021lora}, yet struggle with long-horizon consistency due to limited reasoning data~\cite{guo2024deepseekmathpushinglimitsmathematical}. Reinforcement learning paradigms address these limitations through Direct Preference Optimization (DPO)~\cite{rafailov2024dpo} and Group Relative Policy Optimization (GRPO)~\cite{guo2025deepseekmathgrpo}, which demonstrate significant improvements in chain-of-thought generation for visual domains~\cite{zhang2023multimodalcot,chen2024measuring,qian2024visual}. Hybrid strategies combine SFT with RL through preference learning~\cite{sun2024llava-rlhf}, yet systematic explorations across diverse data sources remain limited~\cite{wang2024reasoning,li2024enhancing}. 
We further explore hybrid optimization schedules beyond standard sequential 
integration (SFT$\rightarrow$GRPO)~\cite{sun2024llava-rlhf}, introducing an alternating 
strategy (SFT$\leftrightarrow$GRPO) that jointly evolves linguistic fluency and reward 
alignment. 


\section{Method}


\begin{figure*}[t]
  \centering
  \includegraphics[width=\linewidth]{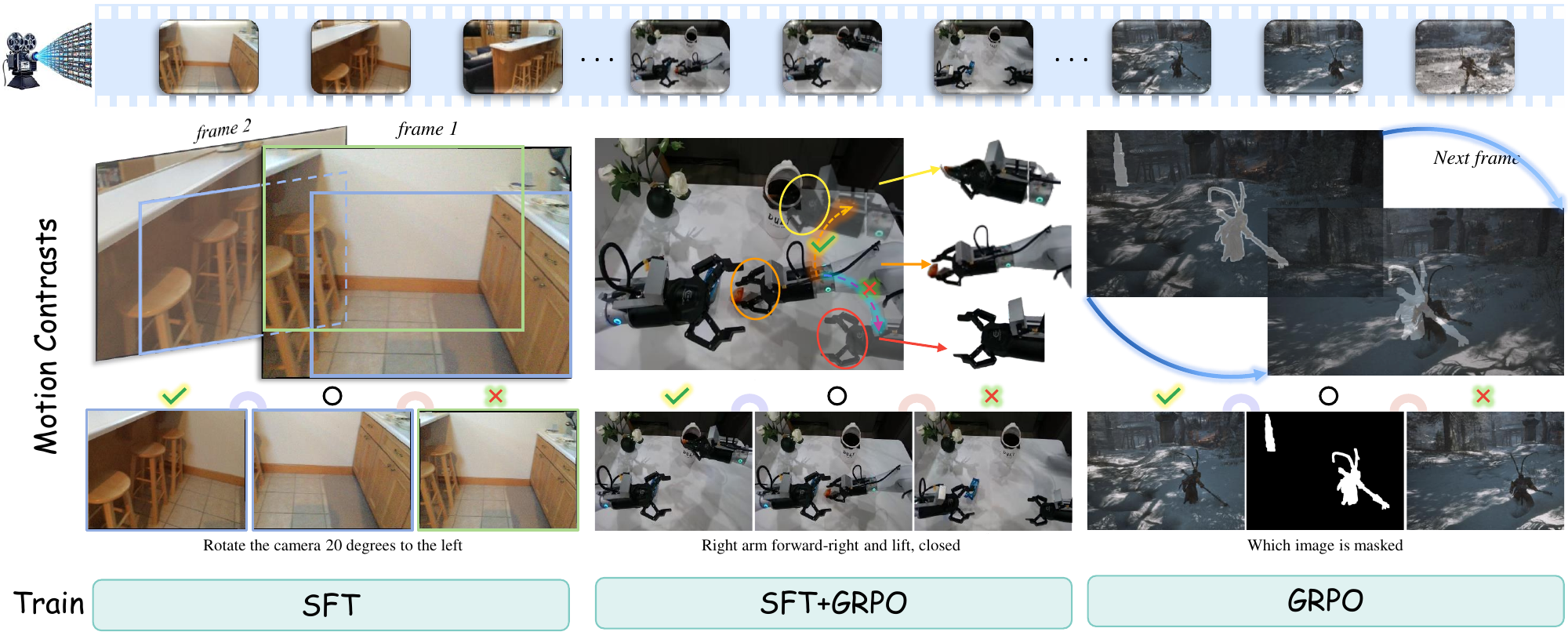}
  \caption{
  \textbf{Overview of the Triplet Motion Contrasts pipeline.}
  Raw videos and meta‑annotations like camera parameters, are processed with rule‑based operations to construct motion‑contrast triplets that encode spatial and temporal changes.
The figure shows representative cases, including camera rotation, manipulation, and masked‑frame contrast, as well as the training paradigms (SFT,SFT+GRPO,GRPO)}
  \label{fig:construction}
\end{figure*}

\subsection{ReMoT-16k Data Construction}
\label{sec:data}

Existing video datasets predominantly provide clip-level captions or static scene descriptions, lacking explicit supervision for distinguishing \emph{physically valid} frame transitions from invalid ones. To address this gap, we propose \method-16k that systematically generates \textbf{motion-contrast triplets} from videos with meta-annotations.

\paragraph{Motion-Contrast Triplets.}

Each triplet consists of $(I_{\text{anchor}}, I_{\text{pos}}, I_{\text{neg}})$, where the anchor-positive pair exhibits a specific motion property $m$ while the anchor-negative pair violates this property despite visual similarity. This design forces models to learn fine-grained motion discrimination rather than relying on superficial visual patterns. Construction requires three critical capabilities: (1) extracting motion properties from frame sequences, (2) constructing motion contrast triplets, and (3) generating VQA formulations.

The most straightforward approach is leveraging vision-language models as unified agents to perform all three stages. We explore this direction using state-of-the-art LVLMs (Qwen-3-VL, Gemini-2.5-Pro), prompting them to analyze video frames and metadata, identify salient motion patterns, and generate structured triplets in JSON format.

However, practical deployment reveals severe limitations. Manual inspection shows 55\% of outputs contain format violations (e.g., mismatched option counts). The approach also incurs prohibitive API costs and time overhead. After quality filtering, we obtain 632 valid triplets. These challenges motivate a more systematic alternative.

\paragraph{Multi-Expert Collaborative Construction.}

Our primary contribution is a multi-expert pipeline where each stage is handled by specialized components operating on structured meta-annotations $\mathcal{A}$:

\begin{enumerate}[leftmargin=*, itemsep=3pt]
\item \textbf{Motion Estimation Experts} ($g: (I_t, I_{t'}, \mathcal{A}) \to m$): Domain-specific extractors parse metadata to derive motion properties. For example, navigation experts compute rigid transformations from $SE(3)$ pose matrices to indicate camera rotation between frames, and manipulation experts extract end-effector trajectories from robot telemetry to indicate robot actions. Each expert outputs compositional properties $m$ encoding motion semantics and attributes.

\item \textbf{Triplet Construction Expert} ($\phi, \mathcal{N}$): This expert first selects salient positive pairs and then synthesizes hard negatives. 
Selection filters for \emph{perceptible yet coherent} transitions via property-specific magnitude thresholds:
\begin{equation}
\phi(I_t, I_{t'}, m) = 
\begin{cases}
(I_{\text{anchor}}, I_{\text{pos}}, m) & \text{if } \|m\| \in \mathcal{T}_m \\
\text{None} & \text{otherwise}
\end{cases}
\end{equation}
where $\mathcal{T}_m$ balances discriminability and coherence (e.g., camera rotation in [10°, 50°]).
Negative generation employs property-conditioned synthesis:
\begin{equation}
\mathcal{N}(I_{\text{anchor}}, I_{\text{pos}}, m) = 
\begin{cases}
\mathcal{T}_{\text{geo}}(I_{\text{pos}}, m)  \\
\mathcal{R}(V, \bar{m}, \text{sim}(I_{\text{pos}}, \cdot)) 
\end{cases}
\end{equation}
where $\bar{m}$ denotes reversed/conflicting attributes. Geometric synthesis $\mathcal{T}_{\text{geo}}$ applies spatial transformations simulating opposite motion, while retrieval $\mathcal{R}$ searches video $V$ for visually similar frames with mismatched properties.

\item \textbf{VQA Formulation Expert}: Given constructed triplets, this expert designs comprehensive question-answer pairs that probe motion understanding from multiple perspectives. Rather than single isolated questions, each triplet $(I_{\text{anchor}}, I_{\text{pos}}, I_{\text{neg}})$ is interrogated through \emph{multi-question reasoning chains} that examine motion properties at different granularities and logical dependencies.
Question design spans diverse formats, such as multiple-choice (selecting valid transitions among options), true/false judgments (verifying motion descriptions), fill-in-the-blank (completing property attributes), and comparative reasoning (relating pairwise motions across all frames). 
\end{enumerate}
Complete descriptions of expert implementations, domain-specific instantiation details, data statistics, and qualitative examples are provided in the supplementary material. 

\paragraph{ReMoT-16k-Test Benchmark.}

To enable rigorous evaluation of fine-grained spatiao-temporal reasoning, we construct \textbf{ReMoT-16k-Test} by combining samples from both construction pipelines. We randomly sample 500 triplets from the multi-expert pipeline as held-out test data, ensuring coverage across all domains: navigation ($\sim$50), robotic manipulation ($\sim$250), others ($\sim$200). We then integrate the 100 VLM-generated triplets covering object tracking and game scenarios, forming a final benchmark of 600 evaluation triplets, with 1776 total evaluation questions. The benchmark exhibits high visual hardness.

\subsection{Training Variants}
\label{sec:training}

Based on the motion–contrast corpus described in Section~\ref{sec:data}, 
we systematically evaluate different optimization paradigms under a unified framework, 
covering supervised fine–tuning (SFT), reinforcement learning (RL), and their hybrid combinations.
All variants use identical training data and comparable configurations.
As illustrated in Fig.~\ref{fig:construction}, three regimes are investigated: 
pure SFT, pure GRPO, and hybrid forms where the two are combined either sequentially or alternately.
 
\noindent\textbf{Base Architecture.}
We choose Qwen3-VL-4B-Thinking as our base model, as it is one of the most capable open--source vision language models for spatio-temporal reasoning.
The 4B variant provides the best balance between accuracy and computational feasibility.
We keep the \emph{Thinking} mode to retain its intrinsic CoT capability 
rather than re-learning reasoning from scratch.
Each sample is formatted as
\begin{center}
\vspace {-3mm}
\texttt{reasoning <think> <answer>ans</answer>}
\vspace {-3mm}
\end{center}
and the loss is computed solely on tokens within \texttt{<answer>} for effective supervision of factual results.
For reinforcement learning, we adopt \emph{Group Relative Policy Optimization (GRPO)}~\cite{grpo}, 
the state--of--the--art paradigm.
Given query~$q$ and $G$ sampled responses~$\{o_i\}_{i=1}^{G}$ from the old policy $\pi_{\theta_{\mathrm{old}}}$, 
the objective is:
\begin{equation}
\begin{aligned}
J(\theta) =
&\;\mathbb{E}_{q,\{o_i\}}\!
\Bigg[
\frac{1}{G}\!
\sum_{i=1}^{G}\!
\min\!\Big(
r_i \hat{A}_i,
\text{clip}(r_i,1\!-\!\varepsilon, 1\!+\!\varepsilon)\hat{A}_i
\Big) \\ 
& - \beta D_{\mathrm{KL}}(\pi_\theta\!\|\!\pi_{\text{ref}})
\Bigg],
\label{eq:grpo}
\end{aligned}
\end{equation}
where $r_i = \pi_\theta(o_i|q) / \pi_{\theta_{\mathrm{old}}}(o_i|q)$ is the probability ratio.
The group-normalized advantages $\hat{A}_i$ are computed as:
\begin{equation}
\hat{A}_i = \frac{R_i - \bar{R}}{\sigma(\{R_j\}_{j=1}^{G})}, \quad \text{where} \quad \bar{R} = \frac{1}{G}\sum_{j=1}^{G} R_j,
\label{eq:advantage}
\end{equation}
with $\sigma(\cdot)$ denoting standard deviation and $R_i$ the total reward for response $o_i$.

\vspace{3pt}

\noindent\textbf{CoT Length Regularization.}
The base model tends to generate excessively verbose and repetitive reasoning traces, 
leading to slower inference and increased memory consumption during training, as shown in Fig.\ref{fig:scaling}.
To address this, we introduce a length penalty into the reward:
\begin{equation}
R_{\text{length}}(o_i) = -\max(0, |o_i^{\text{think}}| - L_{\text{target}}),
\label{eq:cot_reg}
\end{equation}
where $|o_i^{\text{think}}|$ denotes the thinking token count of response $o_i$.
This discourages unnecessarily long chains while preserving reasoning quality.

\vspace{3pt}
\noindent\textbf{Logical Consistency Refinement.}
After baseline training, we analyze failure cases and find that 31.4\% of errors exhibit \emph{logical inconsistency}.
For queries with interdependent questions, models sometimes produce contradictory answers.
For instance, when comparing object lengths across three images, 
a model might claim $L_1 < L_2$, $L_2 < L_3$, yet $L_3 < L_1$ that violates transitivity.
Such contradictions, detectable without ground truth, expose the model's failure to maintain coherent cross-image reasoning.

We formalize a logic verification reward:
\begin{equation}
\label{eq:logicreward}
R_{\text{logic}}(o) = 
\begin{cases}
+1 & \text{answers satisfy logic} \\
-1 & \text{answers contain contradictions} \\
\phantom{+}0 & \text{no verifiable relation exists}
\end{cases}
\end{equation}
The checker applies transitivity to extracted relational symbols.

\vspace{3pt}
\noindent\textbf{Composite Reward Design.}
We integrate all reward components:
\begin{equation}
R_i = R_{\text{task}}(q, o_i) + \lambda_1 \cdot R_{\text{logic}}(o_i) + \lambda_2 \cdot R_{\text{length}}(o_i),
\label{eq:reward_composite}
\end{equation}
where $R_{\text{task}}$ is the task-specific accuracy reward, and $\lambda_1, \lambda_2$ control the strength of logical and length supervision.
These rewards influence the advantages $\hat{A}_i$ in Eq.~(\ref{eq:grpo}), guiding the policy without altering the GRPO objective structure.

\vspace{3pt}
\noindent\textbf{Hybrid Optimization.}
Beyond single-paradigm training, 
we design two integration schemes:
(1)~\emph{Sequential hybrid} (SFT$\rightarrow$GRPO), 
where SFT first provides a stable initialization before the entire policy switches to GRPO refinement; and
(2)~\emph{Alternating hybrid} (SFT$\leftrightarrow$GRPO), 
where SFT and GRPO steps alternate every few updates, allowing linguistic alignment and reward alignment to evolve jointly.  
The alternating procedure is summarized in Algorithm~\ref{alg:hybrid}.

\begin{algorithm}[tb]
   \caption{SFT-GRPO with Composite Rewards}
   \label{alg:hybrid}
\begin{algorithmic}[1]
   \STATE \textbf{Input:} 
   Dataset $\mathcal{D}$, policy $\pi_\theta$, old policy $\pi_{\theta_{\mathrm{old}}}$, 
reference $\pi_{\text{ref}}$,
   max steps $T_{\max}$, group size $G$, clip ratio $\varepsilon$, learning rate $\eta$,
   Phase lengths $(K_{\mathrm{SFT}}, K_{\mathrm{GRPO}})$, weights $\lambda_1, \lambda_2$
   \STATE \textbf{Output:} Optimized parameters $\theta$
   \FOR{$t = 1$ \textbf{to} $T_{\max}$}
      \IF{$(t~\bmod~(K_{\mathrm{SFT}} + K_{\mathrm{GRPO}})) < K_{\mathrm{SFT}}$}
         \STATE Sample ground-truth pair $(q, y^*) \sim \mathcal{D}$
         \STATE Compute cross-entropy loss:
         \STATE \hspace{2em} $\mathcal{L}_{\mathrm{SFT}} = -\!\sum_{u \in \texttt{<answer>}}\! \log \pi_\theta(y_u | q)$
         \STATE Update: $\theta \leftarrow \theta - \eta \nabla_\theta \mathcal{L}_{\mathrm{SFT}}$
      \ELSE
        \STATE Sample prompt $q \sim \mathcal{D}$
        \STATE Generate group responses: $\{o_i\}_{i=1}^{G} \sim \pi_{\theta_{\mathrm{old}}}(\cdot | q)$
        \FOR{$i = 1$ \textbf{to} $G$}
           \STATE Evaluate task reward: $R_{\text{task}}(q, o_i)$
           \STATE Evaluate logical reward: $R_{\text{logic}}(o_i)$ via Eq.~(\ref{eq:logicreward})
           \STATE Evaluate length penalty: $R_{\text{length}}(o_i)$ via Eq.~(\ref{eq:cot_reg})
           \STATE Compute $R_i$ via Eq.~(\ref{eq:reward_composite})
        \ENDFOR
        \STATE Compute group baseline: $\bar{R} = \frac{1}{G}\sum_{i=1}^{G} R_i$
        \STATE Compute advantages: $\hat{A}_i = \frac{R_i - \bar{R}}{\sigma(\{R_i\})}$
        \FOR{$i = 1$ \textbf{to} $G$}
           \STATE Compute importance ratio: $r_i = \frac{\pi_\theta(o_i | q)}{\pi_{\theta_{\mathrm{old}}}(o_i | q)}$
        \ENDFOR
        \STATE Compute objective $J(\theta)$ via Eq.~(\ref{eq:grpo}) 
        \STATE Update: $\theta \leftarrow \theta - \eta \nabla_\theta J(\theta)$
        \STATE Sync old policy: $\pi_{\theta_{\mathrm{old}}} \leftarrow \pi_\theta$
      \ENDIF
   \ENDFOR
\end{algorithmic}
\end{algorithm}

\section{Experiment}

\definecolor{midgray}{gray}{0.95}

\begin{table*}[t]
\centering
\small
\caption{\textbf{Overall and partial accuracies (\%) across capability groups on ReMoT-16k-Test.}
We evaluate models across three fine-grained motion reasoning capabilities: \emph{Navigation}, \emph{Perceptual Grounding}, and \emph{Manipulation}. The rightmost columns show macro-averaged performance across all tasks. Rows compare baseline VLMs against our trained variants: GRPO, SFT$\rightarrow$GRPO (sequential training), and SFT$\leftrightarrow$GRPO (alternating training on held-out data highlighted in gray).}
\setlength{\tabcolsep}{3.8pt}
\renewcommand{\arraystretch}{1.15}
\begin{tabular}{l
                |cc c      
                |c c c     
                |cc cc cc  
                |cc}       
\hline
\multirow{3}{*}{\textbf{Model}} 
& \multicolumn{3}{c|}{\textbf{Navigation}} 
& \multicolumn{3}{c|}{\textbf{Perception}} 
& \multicolumn{6}{c|}{\textbf{Manipulation}} 
& \multicolumn{2}{c}{\textbf{Avg.}} \\
\cline{2-15}
& \multicolumn{2}{c}{Camera} 
& \multicolumn{1}{c|}{Rel‑Pos} 
& \multicolumn{1}{c}{Grounding} 
& \multicolumn{1}{c}{Counting}
& \multicolumn{1}{c|}{} 
& \multicolumn{2}{c}{Gripper‑Move} 
& \multicolumn{2}{c}{Gripper‑State} 
& \multicolumn{2}{c|}{Composite} 
&  &  \\
\cline{2-15}
& Ov. & Par. & Ov.
& Ov. & Ov. & 
& Ov. & Par. & Ov. & Par. & Ov. & Par.
& Ov. & Par. \\
\hline
Qwen2.5‑VL‑7B~\cite{qwen2vl}   & 4.8 & 34.7 & 0.0 & 23.9 & 0.0 & & 4.0 & 36.6 & 8.1 & 31.7 & 0.0 & 16.7 & 5.1 & 25.4 \\
Qwen3‑VL‑CoT-4B \cite{qwen3vl}   & 2.4 & 25.9 & 22.5 & 35.8 & 79.0 & & 15.3 & 46.2 & 3.2 & 35.5 & 4.8 & 22.0 & 20.7 & 38.9 \\
InternVL3‑2B \cite{chen2024internvl} & 1.6 & 21.5 & 20.0 & 31.3 & 60.5 & & 0.8 & 29.8 & 6.5 & 35.0 & 0.0 & 17.7 & 14.9 & 29.3 \\
InternVL3‑8B \cite{chen2024internvl} & 2.8 & 23.8 & 15.0 & 30.6 & 51.7 & & 1.6 & 28.5 & 8.1 & 38.7 & 0.0 & 18.3 & 12.2 & 28.9 \\
LLaVA‑One-Vision \cite{llavaonevision} & 2.0 & 30.4 & 17.5 & 32.9 & 0.0 & & 10.5 & 37.6 & 21.0 & 47.3 & 0.0 & 14.8 & 9.7 & 27.9 \\
\hline
GRPO & \textbf{27.0} & \textbf{62.4} & 24.3
             & 44.3 & \textbf{94.1} & 
             & 54.5 & 74.3 & 27.3 & 44.7
             & 61.3 & 79.2 & 33.6 & 61.6 \\

SFT$\rightarrow$GRPO        & 26.6 & 62.1 & 22.5 & 35.8 & 82.5 & & 57.3 & 78.8 & 32.3 & 60.2 & 62.9 & 82.3 & 35.0 & 63.3 \\
\rowcolor[gray]{0.9}
SFT$\leftrightarrow$GRPO & 21.4 & 61.2 & \textbf{26.7}
             & \textbf{46.7} & 82.7 & 
             & \textbf{68.6} & \textbf{82.0} & \textbf{45.2} & \textbf{65.1}
             & \textbf{69.4} & \textbf{86.6} & \textbf{38.0} & \textbf{64.0} \\
\hline
\end{tabular}
\label{tab:main}
\end{table*}

\subsection{Experimental Setup}

\textbf{Hyper‑parameters.}
We use Qwen3‑VL‑4B‑Thinking \cite{qwen3vl} as our base model. 
During SFT, we use a batch size of~16,
learning rate~$8\!\times\!10^{-5}$, and apply AdamW optimizer with
a cosine decay schedule.
For reinforcement learning, we follow the GRPO~\cite{grpo} scheme and use four rewards:
a \emph{format reward} that encourages valid answer syntax, 
an \emph{accuracy reward} based on correctness against ground truth,
a \emph{conciseness reward} that penalizes excessively long reasoning chains to regulate CoT length,
and a \emph{logic consistency reward} that promotes self-consistent reasoning throughout the inference process.
weighted at~$3.5{:}3.5{:}1.3{:}1.7$.  
Notably, the \emph{format reward} is an internal reward function provided by the ms‑swift library rather than a custom‑designed module; therefore, we do not discuss its implementation details in this paper.
A KL regularization term with coefficient~$0.01$ constrains reward drift from the SFT distribution.
The RL stage uses a batch size of~16 and generates 4~rollouts per sample.
Each training run lasts 2 epochs on \(8\times\)~A800~GPUs using mixed precision.

\noindent\textbf{Evaluation protocols.} We follow the default inference configuration of Qwen3‑VL \cite{qwen3vl}
and evaluate using the \texttt{VLMEvalKit} toolkit~\cite{duan2024vlmevalkit}.
For all benchmarks,
we unify the prompting format and answer extraction rules.
Each question is decoded through the model’s reasoning head
and parsed from the final \texttt{<answer>} token.
Accuracy is computed under both \emph{overall} and \emph{partial} criteria.

\begin{table}[t]
\centering
\small 
\caption{\textbf{Effect of training data composition.}
All results are reported in percentage accuracy on our spatio–temporal benchmark.}
\setlength{\tabcolsep}{6pt}
\renewcommand{\arraystretch}{1.1}
\begin{tabular}{lcc}
\hline
\textbf{Training data} & \textbf{Overall Acc.} & \textbf{Partial Acc.} \\
\hline
Without training & 20.7 & 38.9 \\
Manipulation only & 23.9 & 46.7 \\
+ Navigation & 32.4 & 57.6 \\
+ Simulation  & 38.0 & 64.0 \\
\hline
\end{tabular}
\label{tab:data_comp}
\end{table}


\begin{figure}[t]
\begin{center}
    \includegraphics[width=0.4\textwidth]{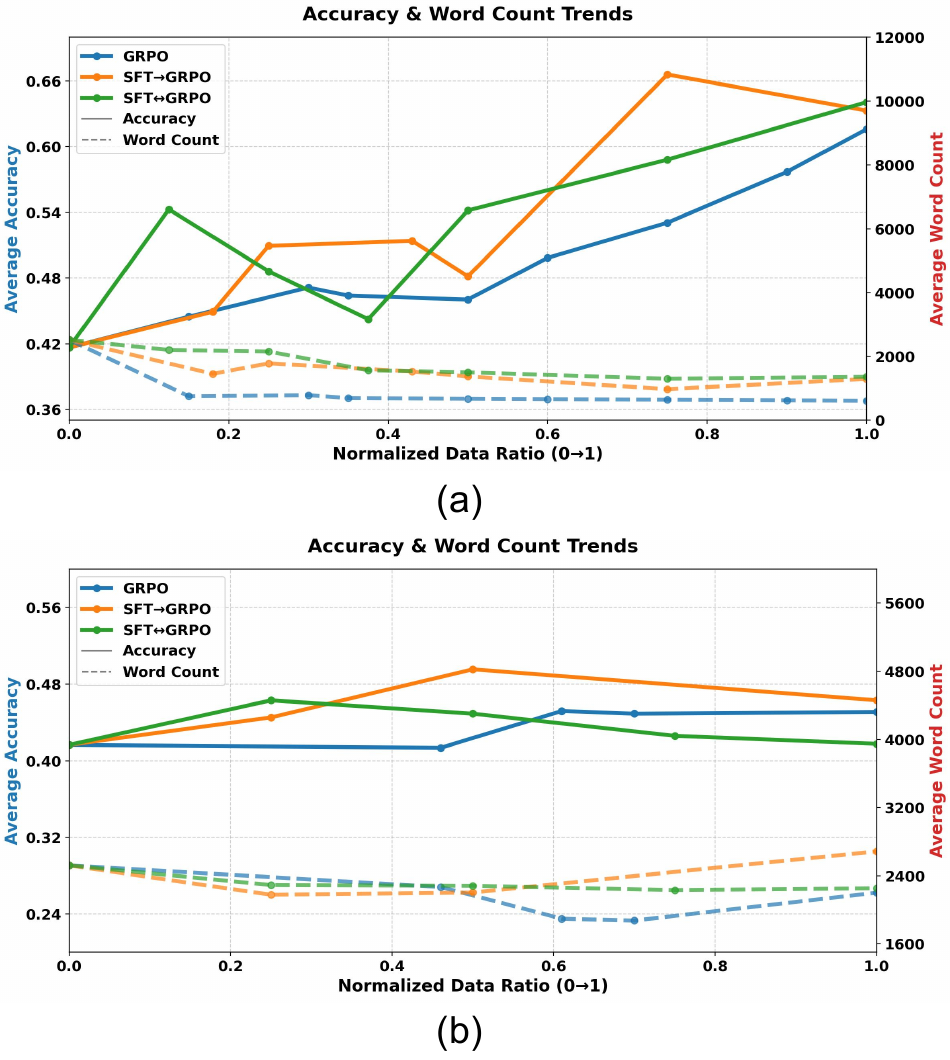}
\end{center}
\vspace{-6pt}
\caption{\textbf{Data scaling analysis across construction pipelines.}
\textbf{(a)} Our multi-expert pipeline shows smooth scaling with GRPO reaching 0.61 and cross-validation variants peaking at 0.64--0.66. 
\textbf{(b)} VLM-generated data exhibits volatile scaling and lower ceiling ($\sim$0.49).
Dashed lines show average word counts, which decline as models learn more concise reasoning with increased data.
}
\vspace{-6pt}
\label{fig:scaling}
\end{figure}

\begin{table}[t]
\centering
\small 
\caption{\textbf{Effect of logic reward decoupling.}
We ablate the contribution of explicit logic consistency reward to model performance. Logic. measures the proportion of responses with consistency.}
\setlength{\tabcolsep}{8pt}
\renewcommand{\arraystretch}{1.1}
\begin{tabular}{lccc}
\hline
\textbf{Method} & \textbf{Ov.} & \textbf{Par.} & \textbf{Logic.} \\
\hline
Qwen3-VL-4B-CoT & 16.2 & 39.6 & 46.6 \\
\quad + GRPO w/o logic reward & 68.6 & 77.3 & 98.6 \\
\quad + GRPO w/ logic reward & \textbf{78.0} & \textbf{81.3} & \textbf{99.3} \\
\hline
\end{tabular}
\label{tab:logic_reward}
\end{table}

\begin{table*}[t]
\centering
\small
\setlength{\tabcolsep}{4.5pt}
\renewcommand{\arraystretch}{1.1}

\begin{tabular}{lccccccc}
\toprule
\multirow{2}{*}{\textbf{Models}} 
& \multicolumn{3}{c}{\textbf{Spatial \& Temporal Benchmarks}} 
& \multicolumn{4}{c}{\textbf{General Multimodal Benchmarks}} \\
\cmidrule(lr){2-4}\cmidrule(lr){5-8}
& \textbf{VLM2~\cite{vlm2bench}} 
& \textbf{VSI~\cite{vsibench}} 
& \textbf{MMSI~\cite{mmsibench}} 
& \textbf{BLINK~\cite{blink}} 
& \textbf{MUIR~\cite{muirbench}} 
& \textbf{MMStar~\cite{MMStar}} 
& \textbf{MMMU~\cite{yue2024mmmu}} \\
\midrule


\rowcolor{line-blue}\textbf{Proprietary} & & & & & & & \\


Gemini‑2.5‑Pro~\cite{gemini2.5}
& 54.2 & 53.6 & 36.9 & 59.1 & 77.2 & 73.6 & 74.7 \\



GPT-4o~\cite{gpt4o}
    & \cellcolor{bestcolor}{60.4} & 42.5 & 30.3 & 59.0 & 68.0 & 70.2 & 70.7 \\

GPT-5~\cite{gpt5}
& \textbf{-} & \cellcolor{bestcolor}{\textbf{55.0}} & \cellcolor{bestcolor}{\textbf{41.8}} 
& \cellcolor{bestcolor}{\textbf{57.9}} & \cellcolor{bestcolor}{\textbf{77.5}} & \cellcolor{bestcolor}{\textbf{75.7}} & \cellcolor{bestcolor}{\textbf{81.8}} \\

\midrule
\rowcolor{line-blue}\textbf{Open‑source} & & & & & & & \\

Qwen2.5‑VL‑7B~\cite{qwen2.5vl}
& 45.1 & 33.0 & 25.9 & 56.4 & 59.6 & 64.1 & 58.0 \\
InternVL2.5‑8B~\cite{chen2024internvl} & 55.4 & 46.6 & \cellcolor{secondcolor}{\textbf{28.7}} & 56.6 & 55.0 & 61.5 & 51.2 \\


LLaVA‑Next‑7B~\cite{llavanext-video} 
& 44.3 & 35.6 & 24.5 & 51.3 & - & 45.3 & 41.7 \\


Qwen3‑VL‑4B-CoT~\cite{qwen3vl}
& 66.4 & 55.2 & 26.8 & 59.5 & \cellcolor{secondcolor}{\textbf{73.8}} & 68.4 & 70.8 \\

Qwen3‑VL-30B-CoT~\cite{qwen3vl}
& \cellcolor{secondcolor}{\textbf{68.2}} & \cellcolor{secondcolor}{\textbf{56.1}} & 28.5 & \cellcolor{bestcolor}{\textbf{65.4}} & \cellcolor{bestcolor}{\textbf{77.6}} & \cellcolor{bestcolor}{\textbf{75.5}} & \cellcolor{bestcolor}{\textbf{76.0}} \\

\textbf{ReMoT-4b-CoT(Ours)} & \cellcolor{bestcolor}{\textbf{70.0}} & \cellcolor{bestcolor}{\textbf{58.8}} & \cellcolor{bestcolor}{\textbf{30.8}} & \cellcolor{secondcolor}{\textbf{62.2}} & 72.8 & \cellcolor{secondcolor}{\textbf{70.4}} & \cellcolor{secondcolor}{\textbf{71.4}} \\

\rowcolor[gray]{0.99}
$\Delta$ \textbf{Improvement} &
\textcolor{ForestGreen}{+3.6} &
\textcolor{ForestGreen}{+3.6} &
\textcolor{ForestGreen}{+4.0} &
\textcolor{ForestGreen}{+2.7} &
‑1.0 &
\textcolor{ForestGreen}{+2.0} &
\textcolor{ForestGreen}{+0.6} \\


\bottomrule
\end{tabular}

\vspace{-2pt}
\caption{\textbf{Evaluation on other VLM benchmarks.}
The first three benchmarks focus on spatio–temporal reasoning, and the latter four evaluate general multimodal understanding.
All scores are reported in accuracy (\%).
Best (\colorbox{bestcolor}{dark purple}) and second‑best (\colorbox{secondcolor}{light purple}) results are highlighted within each group.}
\label{tab:other_benchmarks}
\end{table*}

\subsection{Main Results}
\noindent\textbf{Evaluation metrics.}
Our \method-16k-test benchmark focuses on measuring fine–grained spatial–temporal reasoning consistency.
Each sample in the benchmark contains multiple sub‑questions that jointly evaluate different aspects of reasoning.
For every sample, we define two complementary metrics:
(i)~\emph{Overall Accuracy}, which marks a sample as correct only when all its sub‑questions are answered correctly—that is, any incorrect sub‑question results in the entire sample being scored as wrong; and
(ii)~\emph{Partial Accuracy}, which assigns a proportional score based on the ratio of correctly answered sub‑questions within each sample, allowing partial credit when the reasoning is only partially correct.
The former emphasizes integral reasoning and cross‑frame consistency,
while the latter reflects localized correctness under partial understanding.

\noindent\textbf{Quantitative results.}
Table~\ref{tab:main} presents comprehensive results across three motion reasoning capabilities.
State-of-the-art VLMs achieve below 15\% overall accuracy with near-zero performance on multi-step tasks (Composite Manipulation: 0.0\%–4.8\%), while even reasoning-enhanced Qwen3-VL-CoT shows limited improvement (20.7\%).
Notably, alternating \textbf{SFT--GRPO} training achieves 38.0\% on \emph{Ov} and 64.0\% on \emph{Par}, representing relative gains of \textbf{+18.7\%} and \textbf{+25.1\%} over the base model. 

\subsection{Ablation and Comparative Studies}
\label{sec:ablation}


\noindent\textbf{Training Data Composition.}
Table~\ref{tab:data_comp} shows incremental improvements from different task types.
Starting from the Qwen3-VL baseline (20.7\%), manipulation data provides modest gains (+3.25\%), while adding navigation data brings substantial improvement to 32.35\% (+8.4\%), indicating spatial relation reasoning is crucial.
Incorporating simulation data further achieves 38.0\% (+5.65\%).

\noindent\textbf{Logic Reward Decoupling.} 
Table~\ref{tab:logic_reward} validates our explicit logic supervision on the Manipulation subset.
The logic‑ability test data are mainly drawn from the \emph{Manipulation} subset of our benchmark, which specifically measures reasoning consistency.
Therefore, the base model performance here may differ slightly from the overall results reported in Table~\ref{tab:main}.
We compare three settings: The base model shows poor logic correctness (46.62\%), GRPO without logic reward improves this to 98.6\% but achieves limited accuracy (68.6\%), while our decoupled logic reward reaches 99.3\% logic correctness and 78.0\% accuracy (+10.6\%).
This strong correlation demonstrates that maintaining self-consistent reasoning chains is crucial for performance improvement.

\noindent\textbf{Data Scaling Analysis.}
Figure~\ref{fig:scaling} compares scaling behaviors between pipelines.
Our multi-expert construction (Figure~\ref{fig:scaling}a) exhibits smooth scaling from 0.42 to 0.66 with no saturation, while VLM-generated data (Figure~\ref{fig:scaling}b) shows volatile fluctuations and plateaus at $\sim$0.49.
This demonstrates that construction quality fundamentally limits achievable performance regardless of scale.
Both pipelines show declining response verbosity as data increases.


\begin{figure*}[t]
  \centering
  \includegraphics[width=\linewidth]{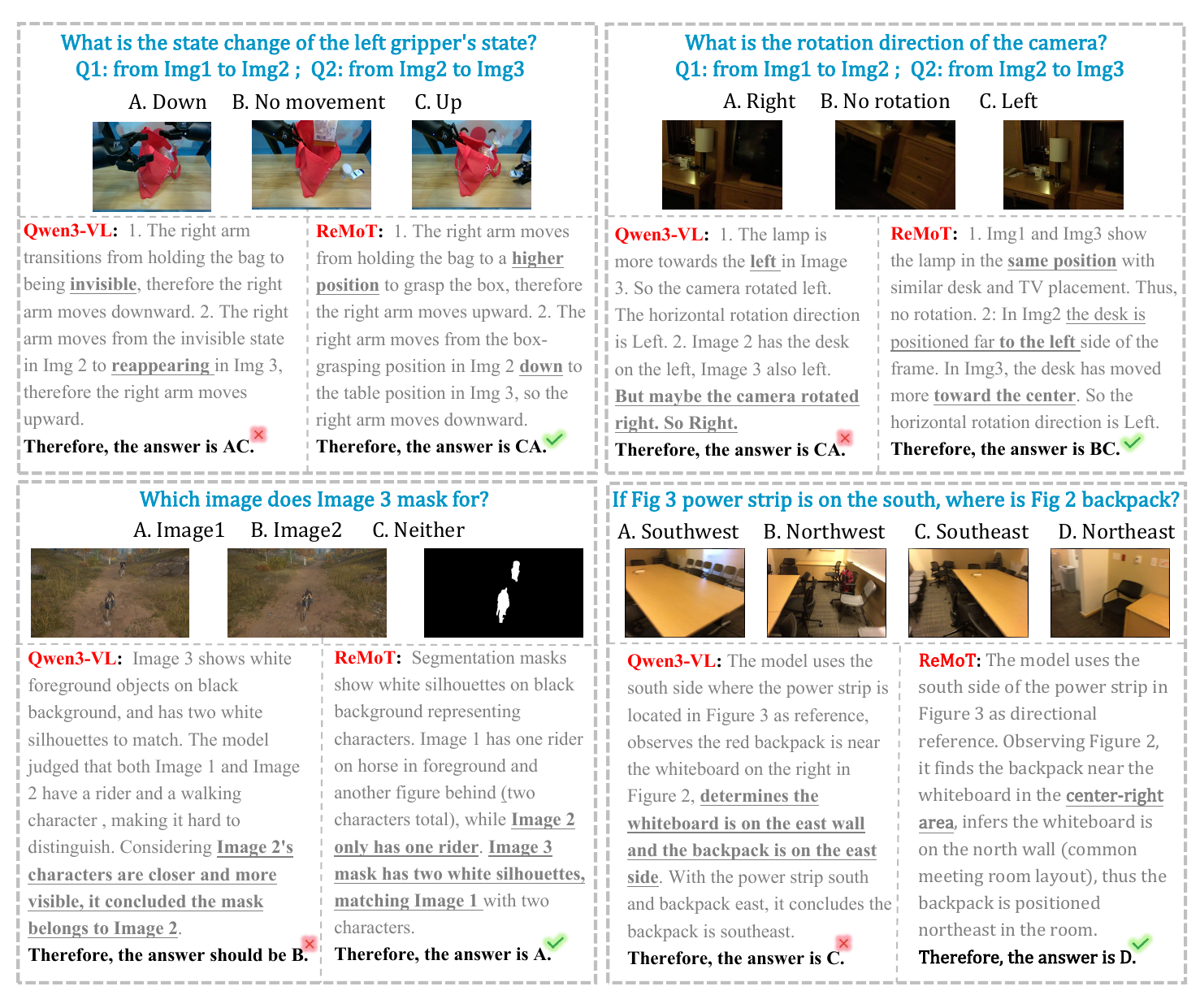}
\caption{\textbf{Visual Comparisons.} 
We compare Qwen3-VL and ReMoT across four challenging scenarios spanning gripper state transitions, camera movement analysis, object segmentation, and directional spatial reasoning. These tasks require distinguishing subtle motion attributes where visual appearances are highly similar but semantic meanings differ significantly. Qwen3-VL frequently misinterprets ambiguous cases and produces contradictory conclusions (underlined in red), while ReMoT leverages structured reasoning chains (highlighted in green) to accurately resolve fine-grained distinctions by integrating temporal dynamics and spatial relationships.}
\vspace{-9pt}
  \label{fig:QUALITY}
\end{figure*}

\subsection{Evaluation on Other VLM Benchmarks}
\label{sec:other_benchmarks}

To assess generalizability, we evaluate ReMoT on existing benchmarks covering spatial-temporal reasoning (VLM2~\cite{vlm2bench}, VSI~\cite{vsibench}, MMSI~\cite{mmsibench}) and general multimodal understanding (BLINK~\cite{blink}, MUIR~\cite{muirbench}, MMStar~\cite{MMStar}, MMMU~\cite{yue2024mmmu}).
The four general benchmarks are selected because they not only measure broad multimodal comprehension, but also include subsets—such as MMStar and BLINK—that explicitly test spatial–temporal understanding in real‑world scenarios.
Results are shown in Table~\ref{tab:other_benchmarks}.
ReMoT-4B-CoT achieves state-of-the-art performance among all models on spatial-temporal benchmarks, outperforming the 7.5× larger Qwen3-VL-30B-CoT by +1.8\%, +2.7\%, and +2.3\% on VLM2, VSI, and MMSI respectively.
On the four general benchmarks, ReMoT maintains comparable or better performance, indicating that strengthening spatial–temporal reasoning does not compromise general multimodal capability.
Notably, our 4B model matches GPT-4o on spatial-temporal tasks despite being orders of magnitude smaller.
The consistent gains on spatial-temporal benchmarks while preserving general capabilities confirm our expert-driven data construction, decoupled logic rewards, and GRPO optimization successfully enhance core reasoning without catastrophic forgetting.

\subsection{Qualitative Analysis.}
\vspace{-5pt}
We sample representative examples from our benchmark and visualize side-by-side the baseline’s reasoning and ReMoT’s reasoning traces, along with their textual responses.

\vspace{-8pt}
\section{Conclusion}
\label{sec:conclusion}
\vspace{-6pt}
We present ReMoT, a comprehensive framework for fine-grained spatial-temporal reasoning in vision-language models. Through multi-expert collaboration, we construct ReMoT-16K, a dataset explicitly modeling subtle inter-frame motion attributes. Combined with decoupled logic rewards and GRPO optimization, ReMoT achieves state-of-the-art performance on spatial-temporal benchmarks while maintaining competitive results on general multimodal tasks.

\section{Acknowledgement}
\label{sec:Acknowledgement}
This work is supported by the National Natural Science Foundation of China under Grant No.U21B2048 and No.62576224, the Shenzhen Key Technical Projects under Grant CJGJZD20220517141605013, JCYJ20220818101406014 cand JSGG20220831105801004, and Guangdong Provincial Key Laboratory of Computility Microelectronics with Grant 2024B1212010007.

{
    \small
    \bibliographystyle{ieeenat_fullname}
    \bibliography{main}
}

\end{document}